\documentclass[runningheads]{llncs}
\usepackage{float}
\usepackage[caption = false]{subfig}
\usepackage[final]{graphicx}
\usepackage{array}
\usepackage{multirow}
\usepackage{hhline}
\usepackage[width=122mm,left=12mm,paperwidth=146mm,height=193mm,top=12mm,paperheight=217mm]{geometry}
\usepackage{color}
\usepackage{amsmath,amssymb} 

\DeclareMathOperator*{\argmax}{argmax}
\DeclareMathOperator*{\argmin}{argmin}
\newcommand{\planes}{\mathcal{P}}
\newcommand{\motion}{\mathcal{X}}
\newcommand{\plane}{\bar{\mathbf{n}}}
\newcommand{\edge}{\mathcal{E}}
\newcommand{\Rone}{\mathbf{R}}
\newcommand{\Rtwo}{\mathbf{R}^2}
\newcommand{\Rthree}{\mathbf{R}^3}
\newcommand{\Rsix}{\mathbf{R}^6}
\newcommand{\SOthree}{\mathbf{SO}(3)}
\newcommand{\SEthree}{\mathbf{SE}(3)}
\newcommand{\rotation}{\mathbf{R}}
\newcommand{\translation}{\mathbf{t}}
\newcommand{\Pthree}{\mathbf{x}}
\newcommand{\extrinsics}{\mathcal{T}}
\newcommand{\calibration}{\mathbf{K}}
\newcommand{\transform}{\mathcal{T}}

\newcommand\Markauthor[1]{\textsuperscript{#1}}

\begin{document}

\pagestyle{headings}
\mainmatter

\title{A Continuous Optimization Approach\\for Efficient and Accurate Scene Flow} 

\titlerunning{A Continuous Optimization Approach for Scene Flow}

\authorrunning{Z. Lv et al.}

\author{Zhaoyang Lv\Markauthor{1}, Chris Beall\Markauthor{1}, 
	Pablo F. Alcantarilla\Markauthor{3}, 
	Fuxin Li\Markauthor{4}, \\ 
	Zsolt Kira\Markauthor{2}, 
	Frank Dellaert\Markauthor{1} }
\institute{
	\Markauthor{1} Georgia Institute of Technology, Atlanta, US \\
    \email{ \{zlv30,cbeal3\}@gatech.edu dellaert@cc.gatech.edu } \\
    \Markauthor{2} Georgia Tech Research Institute, Atlanta US \\
    \email{ zkira@gatech.edu } \\ 
    \Markauthor{3} iRobot Corporation, London, UK \\
    \email{palcantarilla@irobot.com} \\
    \Markauthor{4} Oregon State University, Corvallis, US \\
    \email{lif@eecs.oregonstate.edu}
} 

\maketitle

\begin{abstract}
We propose a continuous optimization method for solving dense 3D scene flow problems from stereo imagery. As in recent 
work, we represent the dynamic 3D scene as a collection of rigidly moving planar segments. 
The scene flow problem then becomes the joint estimation of pixel-to-segment assignment, 
3D position, normal vector and rigid motion parameters for each segment, leading to a complex and expensive discrete-continuous optimization 
problem. In contrast, we propose a purely continuous formulation which can be solved more efficiently. 
Using a fine superpixel segmentation that is fixed a-priori, we propose a factor graph formulation that decomposes the problem into photometric, geometric, and smoothing constraints. 
We initialize the solution with a novel, high-quality initialization method, then independently refine the geometry and motion of the 
scene, and finally perform a global non-linear refinement using Levenberg-Marquardt.
We evaluate our method in the challenging KITTI Scene Flow 
benchmark, ranking in third position, while being 3 to 30 times faster than the top competitors (x37~\cite{Menze15cvpr} and x3.75~\cite{Vogel15ijcv}). 
\end{abstract}


\section{Introduction}\label{sec:introduction}\label{sec:relatedwork}

\begin{figure}[t]
\begin{center}
 \includegraphics[width=\linewidth]{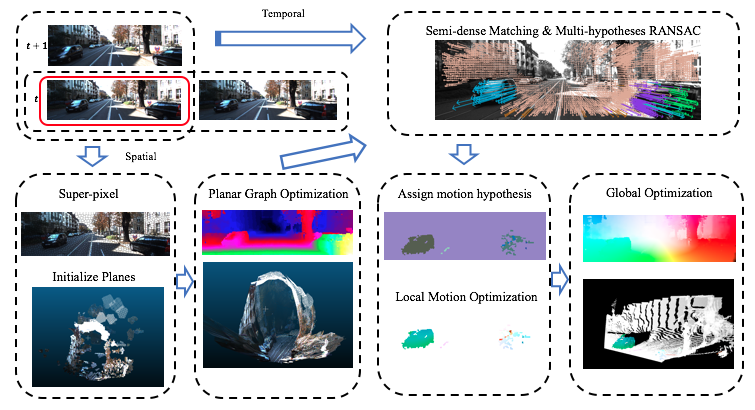}
\end{center}
\caption{An overview of our system: we estimate the 3D scene flow w.r.t. the reference image (the red bounding box), a stereo image pair and a temporal image pair as input. Image annotations show the results at each step. We assign a motion hypothesis to each superpixel as an initialization and optimize the factor graph for more accurate 3D motion. Finally, after global optimization, we show a projected 2D flow map in the reference frame and its 3D scene motion (static background are plotted in white).}
\label{fig:pipeline}
\end{figure}


Understanding the geometry and motion within urban scenes, using either monocular or stereo imagery, is an important problem with increasingly relevant applications such as 
autonomous driving \cite{Rabe10eccv}, urban scene understanding \cite{Rabe10eccv,Pfeiffer10iv,Wedel08eccv}, video analysis 
\cite{Hung13ijcv}, dynamic reconstruction \cite{Pons07ijcv,Newcombe15cvpr}, etc. 
In contrast to separately modeling 3D geometry (stereo) and characterizing 
the movement of 2D pixels in the image (optical flow), the scene flow problem is to characterize the 3D motion 
of points in the scene \cite{Vedula05iccv} (Fig.~\ref{fig:pipeline}). Scene flow in the context of stereo 
sequences was first investigated by Huguet et al. \cite{Huguet07iccv}. Recent work 
\cite{Menze15cvpr,Vogel13iccv,Valgaerts10eccv} has shown that explicitly reasoning about the scene flow can in turn improve both stereo and optical flow estimation.

Early approaches to scene flow ranged from directly estimating 3D displacement from stereo~\cite{Zhang92pami}, using volumetric 
representations \cite{Vedula05iccv,Vedula05pami} in a many-camera setting,  to re-casting the problem as a 2D disparity flow 
\cite{Huguet07iccv,Isard06accv} in motion stereo settings.
A joint optimization is often leveraged to solve an energy model with all spatio-temporal constraints, e.g. 
\cite{Menze15cvpr,Vogel13iccv,Huguet07iccv,Basha12ijcv}, but \cite{Valgaerts10eccv} argues for solving scene and camera motion in an 
alternating fashion. \cite{Wedel11ijcv} claims that a decomposed estimation of disparity and motion field can be advantageous as each step 
can use a different optimization technique to solve the problem more efficiently. A real-time semi-dense scene flow can be achieved without 
loss of accuracy.

However, efficient and accurate estimation of scene flow is still an unsolved problem. Both dense stereo and optical flow are challenging problems in their own right, and reasoning about the 3D scene 
must still cope with an equivalent aperture problem \cite{Vedula05iccv}.
In particular, in scenarios where the scene scale is much larger than 
the stereo camera baseline, scene motion and depth are hardly distinguishable. Finally, when there is significant motion in the scene there 
is a large displacement association problem, an unsolved issue for optical flow algorithms.

Recently, approaches based on a rigid moving planar scene assumption 
have achieved impressive results \cite{Menze15cvpr,Vogel13iccv,Vogel14eccv}. 
In these approaches, the scene is represented 
using planar segments which are assumed to have consistent motion. The scene flow problem is then posed as a discrete-continuous 
optimization problem which associates each pixel with a planar segment, each of which has continuous rigid 3D motion parameters to be 
optimized.
Vogel et al. \cite{Vogel13iccv} view scene flow as a discrete labeling 
problem: assign the best label to each super-pixel plane from a set of moving plane proposals. 
\cite{Vogel14eccv} additionally leverages a temporal sequence to achieve consistency both in depth and motion estimation.
Their approach casts the entire problem into  a discrete optimization problem. 
However,  joint inference in this space is both complex and computationally expensive. 
Menze and Geiger~\cite{Menze15cvpr} partially 
address this by parameter-sharing between multiple planar segments, by assuming the existence of a finite set of moving objects in the 
scene. 
They solve the candidate motion of objects with continuous 
optimization, and use discrete optimization to assign the label of each object to each superpixel. 
However, this assumption does not hold for scenes with non-rigid deformations. Piece-wise continuous planar assumption is not limited to 3D description. \cite{Yang15cvpr} achieves state-of-art optical flow results using planar models.

In contrast to this body of work, we posit that it is better to solve for the scene flow in the continuous domain. We adopt the 
same rigid planar representation as \cite{Vogel13iccv}, but solve it more efficiently with high accuracy. Instead of reasoning about discrete labels, we use a fine 
superpixel segmentation that is fixed a-priori, and utilize a robust nonlinear least-squares approach to cope with occlusion, depth and motion discontinuities in the scene. A central assumption is that once a fine enough superpixel segmentation is used as a priori, there is no need to jointly optimize the superpixel segmentation within the system. The rest of the scene flow problem, being piecewise continuous, can be optimized entirely in continuous domain. A good initialization is obtained by leveraging \textit{DeepMatching} ~\cite{Weinzaepfel13iccv}. We achieve fast inference by using a sparse nonlinear least squares solver and avoid discrete approximation. To utilize Census cost for fast robust cost evalution in continuous optimization, we proposes a differentiable Census-based cost, similar to but not same as the approach in \cite{Vogel13gcpr}.

This work makes the following contributions: first, we propose a factor-graph formulation of the scene flow problem that exposes the 
inherent sparsity of the problem,  and use a state of the art sparse solver that directly optimizes over the manifold representations of the 
continuous unknowns. Compared to the same representation in~\cite{Vogel13iccv}, we achieve better accuracy and faster inference. Second, instead of directly 
solving for all unknowns, we propose a pipeline to decompose geometry and motion estimation. We show that this helps us cope with the highly 
nonlinear nature of the objective function. Finally, as initialization is crucial for nonlinear optimization to succeed, we use the 
DeepMatching algorithm from~\cite{Weinzaepfel13iccv} to obtain a semi-dense set of feature correspondences from which we initialize the 
3D motion of each planar segment. As in~\cite{Menze15cvpr}, we initialize planes from a restricted set of motion hypotheses, but optimize them in 
the continuous domain to cope with non-rigid objects in the scene. 
\section{Scene Flow Analysis} \label{sec:sceneflow}

We follow~\cite{Vogel13iccv} in assuming that our 3D world is composed of locally smooth and rigid objects. 
Such a world can be represented as a set of rigid planes moving in 3D, $\planes={\{\plane, \motion\}}$, with parameters representing the 
plane normal $\plane$ and motion $\motion$. In the ideal case, a slanted plane 
projects back to one or more superpixels in the images, inside of which the appearance and geometry information are locally similar. The 
inverse problem is then to infer the 3D planes (parameters $\plane$ and $\motion$), given the images and a set of pre-computed superpixels.

\begin{description}
\item[3D Plane]
We denote a plane as $\plane$ in 3-space, specified by its normal coordinates in the reference frame. For any 3D point $\Pthree \in 
\Rthree$ on 
$\plane$, the plane equation holds as $\plane^{\top} \Pthree + 1=0$. We choose this parameterization for ease of optimization on its 
manifold (refer to Section \ref{sec: continuous_optimization}.) 
\item[Plane Motion] 
A rigid plane transform $\motion \in \SEthree$ comprising rotation and translation is defined by
\begin{equation}
\motion = \begin{bmatrix}
\rotation & \translation \\
\mathbf{0}  & 1
\end{bmatrix} 
, \rotation \in \SOthree, \translation \in \Rthree
\end{equation}
\item[Superpixel Associations] 
We assume each superpixel $S_i$ is a one-to-one mapping from the reference frame to a 3D plane.
The boundary between adjacent superpixels $S_i$ 
and $S_j$ is defined as $\edge_{i,j} \in \Rtwo$.
\end{description}

\subsection{Transformation induced by moving planes} \label{sec:transformation_induced_by_planes}

For any point $\Pthree$ on $\plane$, its homogeneous representation is $[\Pthree^{\top}, -\plane^{\top} \Pthree]$. From $\Pthree_0$ in the reference frame, its corresponding point $\Pthree_1$ in an observed frame is:

\begin{equation} \label{eq: plane_induced_projection_chain}
\begin{bmatrix}
\Pthree_1 \\
1
\end{bmatrix} =
\begin{bmatrix}
	\rotation_{0}^{1} & \translation_{0}^{1} \\ 
	0 & 1
\end{bmatrix}
\begin{bmatrix}
	\rotation_i & \translation_i \\
	0 & 1
\end{bmatrix}
\begin{bmatrix}
\Pthree_0 \\
-\plane^{T} \Pthree_0
\end{bmatrix}
\end{equation}

\noindent where $[\rotation_{0}^{1}|\translation_{0}^{1}]$ is the transform from reference frame to the observed image frame (referred to as $\mathcal{T}^{1}_{0}$) and $[\rotation_i|\translation_i]$ is the plane motion in the reference frame (referred to as $\motion_i$). Suppose the camera intrinsic matrix as $\calibration$, A homography transform can thus be induced as:

\begin{equation} \label{eq: homography_chain_2}
\begin{aligned}
\mathbf{H}(\planes_i, \transform^{1}_{0}) &= \calibration[\mathbf{A} - \mathbf{a}\plane]\calibration^{-1} \\
\begin{bmatrix}
\mathbf{A} & a\\
0 & 1
\end{bmatrix}
&=
\begin{bmatrix}
	\rotation_{0}^{1} & \translation_{0}^{1} \\ 
	0 & 1
\end{bmatrix}
\begin{bmatrix}
	\rotation_i & \translation_i \\
	0 & 1
\end{bmatrix}
\end{aligned}
\end{equation}

In stereo frames where planes are static, the homography from reference frame to the right frame is simply:
\begin{equation} \label{eq:homograpy_stereo}
\mathbf{H}(\plane, \transform^{r}_{0}) = \calibration(\rotation^{r}_{0}-\translation^{r}_{0}\plane)\calibration^{-1}
\end{equation}

We will only use $\transform^{r}_{0}$ to represent the transform of reference frame to the other stereo frame, while $\transform^{1}_{0}$ is applicable from reference frame to any other frames, whether the planes are static or moving.


\subsection{A Factor Graph Formulation for Scene Flow} \label{sec:factor_graph}

\begin{figure*}[t]
\begin{center}
\includegraphics[width=0.9\linewidth]{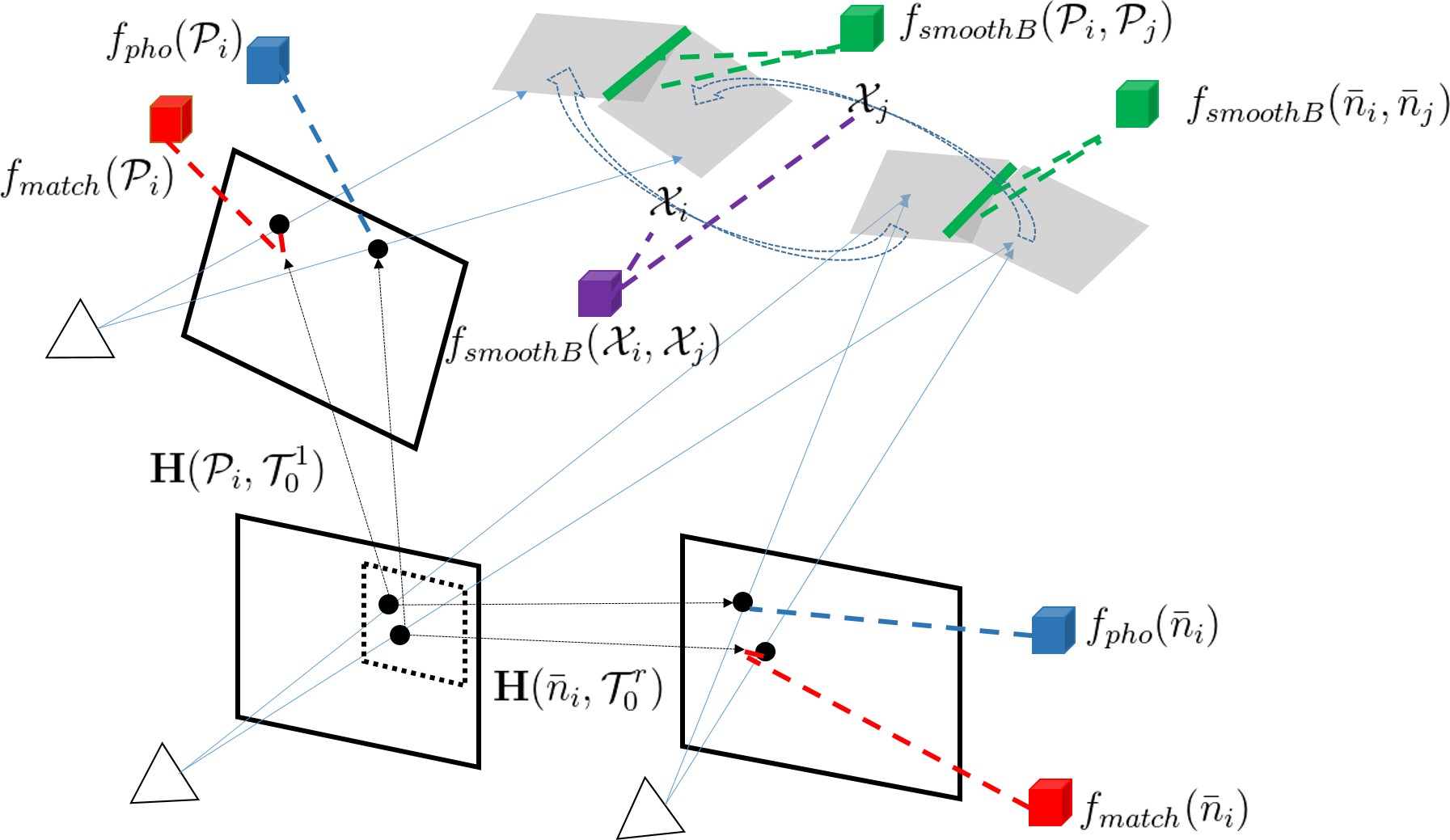}
   \caption{The proposed factor graph for this scene flow problem. The unary factors are set up based on the homography transform relating two pixels, given $\planes$. Binary factors are set up based on locally smooth and rigid assumptions. In this graph, a three-view geometry is used to explain factors for simplicity. Any other views can be constrained by incorporating the same temporal factors in this graph.}
\label{fig:factor_graph}
\end{center}
\end{figure*}


\newcommand{\domain}{\Omega}
For all images $I': \domain \to \Rone$ relative to the reference image $I: \domain \to \Rone$, we want to 
estimate all of the planes $\Theta = \{\plane_{\{1...N\}}, \motion_{\{1...N\}} \}$ observed in $I$. Besides raw image measurements, we also 
assume that a set of sparsely matched point pairs $M \in \Rtwo$ is available. As mentioned above, we assume an a-priori fixed superpixel 
segmentation $S$, along with its boundaries $\edge$. We denote these as our measurements $\mathcal{M} = \{I, I', M, S, 
\edge \}$.

We begin by defining parameters $\theta = \{\plane, \motion\}$, in which $\plane$ and $\motion$ are independent to each other. We also assume  
dependencies only exist between superpixels across common edges. The joint probability distribution of $\Theta$ can then be:

\begin{equation} \label{eq:factor_graph}
\begin{aligned}
\mathbf{P}(\Theta, \mathcal{M}) &\propto \prod_{i \in N}\mathbf{P}(\theta_i |\mathcal{M}) \prod_{j \in N \backslash \{i\}} 
\mathbf{P}(\theta_i, \theta_j | \mathcal{M}) \\
\mathbf{P}(\theta_i | \mathcal{M}) &\propto \mathbf{P}(I', M|\plane_i, \motion_i, S_i, I) \mathbf{P}(\plane_i) \mathbf{P}(\motion_i) \\
\mathbf{P}(\theta_i, \theta_j|\mathcal{M}) &= \mathbf{P}(\plane_i, \plane_j|S_i, S_j, \edge_{i,j}) \mathbf{P}(\motion_i,\motion_j|S_i, S_j, 
\edge_{i,j}),
\end{aligned}
\end{equation} 

Factor graphs (see e.g., \cite{Kschischang01it}) are convenient probabilistic graphical models for formulating the scene flow problem:

\begin{equation}
G(\Theta) = \prod_{i\in N}f_i(\theta_i)\prod_{i,j \in N}f_{ij}(\theta_i, \theta_j),
\end{equation}

Typically $f(\theta_i)$ encodes a prior or a single measurement constraint at unknown $\theta$, and $f_{i,j}$ relate to measurements or 
constraints between $\theta_i, \theta_j$. In this paper, we assume each factor is a least-square error term with Gaussian noises. To fully represent the measurements and constraints in this problem, we will use multiple 
factors for $G(\Theta)$ (see Fig.~\ref{fig:factor_graph}), which will be illustrated below.

\subsubsection*{Unary Factors}

A point $p$, associated with a particular superpixels, can be associated with the homography transform $\mathbf{H}(\planes_i, 
\extrinsics_s)$  w.r.t. its measurements. For a stereo camera, the transformation of a point from one image to the other is simply 
$\mathbf{H}(\plane, \extrinsics_s)$ in Eq.~\ref{eq:homograpy_stereo}. For 
all the pixels $p$ in superpixel $S_i$, their photometric costs given $\planes \{ \plane_i, \motion_i \}$ is described by factor 
$f_{pho}(\planes_i)$:

\begin{equation} \label{eq:census_factor}
f_{pho}(\planes_i) \propto \prod_{p \in S_i} f \big( C(p'), C(\mathbf{H}(\planes_i, \transform^{1}_{0})p \big),
\end{equation}

\noindent where $C(\cdot)$ is the Census descriptor. This descriptor is preferred over intensity error for its robustness against noise and 
edges. Similarly, using the homography transform and with sparse matches we can estimate the geometric error of match $m$ by measuring its 
consistency with the corresponding plane motion: 

\begin{equation} \label{eq:matching_factor}
f_{match}(\planes_i) \propto \prod_{p \in S_i} f \big(p + m , \mathbf{H}(\planes_i, \transform^{1}_{0})p \big),
\end{equation}

\subsubsection*{Pairwise Factors}
The pairwise factors relate the parameters based on their constraints. $f_{smoothB}(\cdot, \cdot)$ describes the locally smooth assumption 
that adjacent planes should share similar boundary connectivity:

\begin{equation} \label{eq:smooth_boundary_plane}
f_{smoothB}(\plane_i, \plane_j) \propto \prod_{p \in \edge_{i,j}} f \big( D^{-1}(\plane_i, p), D^{-1}(\plane_j, p) \big),
\end{equation}

\noindent where $D^{-1}(\plane, p)$ represents the inverse depth of pixel $p$ on $\plane$. This factor describes the distance of points over the 
boundary of two static planes. After plane motion, we expect the boundary to still be connected after the transformation:

\begin{equation} \label{eq:smooth_boundary_motion}
f_{smoothB}(\planes_i, \planes_j) \propto \prod_{p \in \edge_{i,j}} f \big( D^{-1}(\planes_i, p), D^{-1}(\planes_j, p) \big),
\end{equation}

With our piece-wise smooth motion assumption, we also expect that two adjacent superpixels should share similar motion parameters, described by 
$f_{smoothM}$, which is a \emph{Between} operator of $ \SEthree$:

\begin{equation} \label{eq:smooth_motion}
f_{smoothM}(\motion_i, \motion_j) \propto f \big(\motion_i, \motion_j \big).
\end{equation}

Each factor is created as a Gaussian noise model: $f(x;m) = \exp(-\rho(h(x)-m)_\Sigma)$ for unary factor and $f(x_1, x_2) = \exp(-\rho(h_1(x_1) - h_2(x_2))_\Sigma)$ for binary factor. $\rho(\cdot)_\Sigma$ is the Huber robust cost which measures the Mahalanobis norm. It incorporates the noise effect of each factor and down-weights the effect of outliers. Given a decent initialization, this robust kernel helps us to cope with occlusions, depth and motion discontinuities properly.

\subsection{Continuous Optimization of Factor Graph on Manifold} \label{sec: continuous_optimization}

The factor graph in Eq.~\ref{eq:factor_graph} can be estimated via maximum a posteriori (MAP) as a non-linear least square problem, 
and solved with standard non-linear optimization methods. In each step, we linearize all the factors at $\theta =\{\plane_{\theta}, 
\motion_{\theta}\}$. On manifold, the update is a \emph{Retraction} $\mathcal{R}_{\theta}$. The retraction for $\{\plane, \motion\}$ is:

\begin{equation} \label{eq:retraction_detail}
\mathcal{R}_\theta(\delta \plane, \delta \motion) = (\plane+\delta\plane, \motion\text{Exp}(\delta x)), [\delta \plane \in \Rthree, \delta x 
\in \Rsix]
\end{equation}

For $\plane \in \Rthree$, it has the same value of its tangent space at any value $\hat{n}$. This explains our choice of plane 
representation: it is the most convenient for manifold optimization in all of its families in 3-space. For motion in $\SEthree$, the 
retraction is an exponential map.

Although the linearized factor graph can be thought of as a huge matrix, it is actually quite sparse in nature: pairwise factors only exist between adjacent superpixels. Sparse matrix factorization can solve this kind of problem very efficiently. We follow the same sparse matrix factorization which is discussed in detail in \cite{Dellaert06ijrr}. 

\subsection{Continuous Approximation for Census Transform} \label{sec: continuous_census}

\begin{figure}[t] 
\begin{center}
\subfloat{\includegraphics[width=0.40\linewidth]{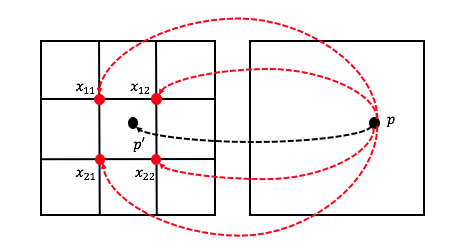}} \ \ 
\subfloat{\includegraphics[width=0.40\linewidth]{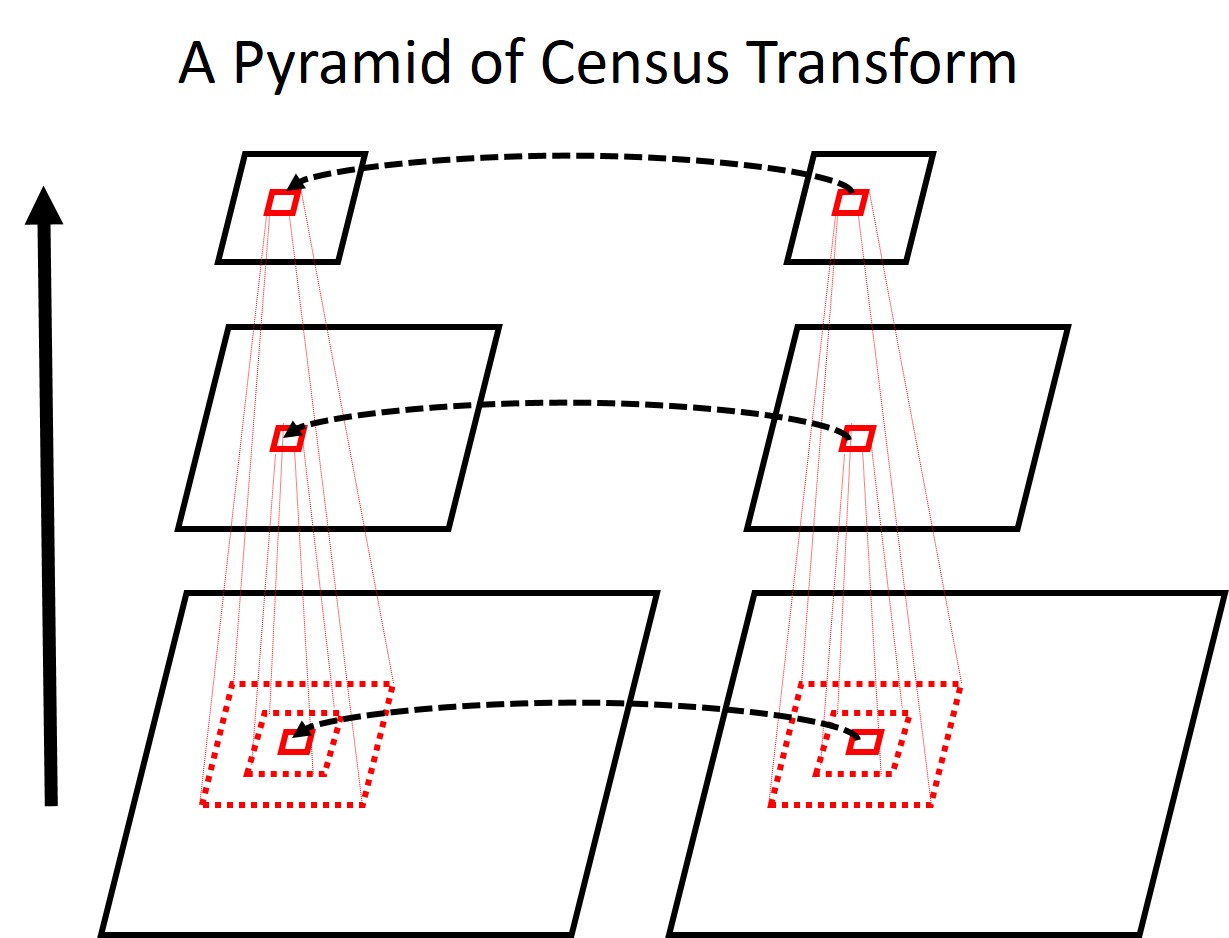}}
\caption{The left figure shows how to use bilinear interpolation to achieve a differentiable cost of Census Transform. In the right figure, 
a census descriptor is extracted at different pyramid levels of the images. When evaluating its distance w.r.t. another pixel, we also use bilinear 
interpolation to evaluate census cost at lower resolution images.}
\label{fig:census}
\end{center}
\end{figure}

In Eq.~\ref{eq:census_factor}, there are two practical issues: first, we cannot get a sub-pixel Census Transform; and second, the Hamming 
distance between the two descriptors is not differentiable. To overcome these problems, we use bilinear interpolated distance as the census 
cost (see Fig.~\ref{fig:census}). The bilinear interpolation equation is differentiable w.r.t. the image coordinate, from which we can approximately get the Jacobian of 
Census Distance w.r.t. to a sub-pixel point. We use a 9x7 size Census, and set up Eq.~\ref{eq:census_factor} over a pyramid of 
images. In evaluation, we will discuss how this process helps us to achieve better convergence purely with a data-cost.

\section{Scene Flow Estimation}\label{sec:pipeline} \label{sec:pipeline_overview} 

The general pipeline of our algorithms consists of five steps (see Fig.~\ref{fig:pipeline}). We summarize each step and provide detailed descriptions in the subsections below.

\noindent{}\textbf{Initialization} We initialize the superpixels for the reference frame. For both of the stereo pairs, we estimate a depth map as priors. The 3D plane is initialized from the depth map using RANSAC.

\noindent{}\textbf{Planar Graph Optimization} We solve the factor graph composed of factors in Eq.~\ref{eq:census_factor}, 
\ref{eq:matching_factor} and~\ref{eq:smooth_boundary_plane}. The result is the estimation of plane geometry parameter $\plane$ w.r.t. 
reference frame.

\noindent{}\textbf{Estimation of Motion Hypotheses} We first estimate a semi-dense matching from reference frame to the next temporal frame and 
associate them with our estimated 3D plane to get a set of 3D features. We use RANSAC to heuristically find a set of motion hypothesis. In 
each RANSAC step, we find the most likely motion hypothesis of Eq.~\ref{eq: homography_chain_2} by minimizing the re-projection errors 
of 3D features in two temporally consecutive frames. A set of motion hypotheses are generated by iterating this process.

\noindent{}\textbf{Local Motion Graph Optimization} We initialize the motion of superpixels from the set of motion hypotheses, framed as a Bayesian 
classification problem. For all of the superpixels assigned to one single motion hypothesis, we estimate both the plane $\plane$ and its motion 
$\motion$, by incorporating factors in Eq.~\ref{eq:census_factor}, \ref{eq:smooth_boundary_motion}, \ref{eq:smooth_motion}.

\noindent{}\textbf{Global Graph Optimization} In this step, the set of all unknowns $\planes$ is estimated globally. All factors from 
Eq.~\ref{eq:census_factor} to \ref{eq:smooth_motion} are used.

\subsection{Initialization} \label{sec:initialiation}
The superpixels in the reference frame are initialized with the sticky-edge superpixels introduced in~\cite{Zitnick14eccv}. Since the urban 
scene is complex in appearance, the initialized superpixel number needs to be large to cope with tiny objects, while too many superpixels 
can cause an under-constrained condition for some plane parameters. Empirically, we find generating 2,000 superpixels is a good balance (refer to our superpixel discussion in supplement materials.)

We use the stereo method proposed in~\cite{Yamaguchi14eccv} to generate the stereo prior, and initialize the 3D planes with a plane-fitting RANSAC algorithm. The plane is initialized as frontal parallel if the RANSAC inlier percentage is below a certain threshold (50\% in our setting), or the plane induces a degenerated homography transform (where the plane is parallel to the camera focal axis). 

We sample robust matches $\mathcal{M}$ from the disparity map, and use it to set up the matching factor in Eq.~\ref{eq:matching_factor}. The samples are selected from the Census Transform which share a maximum distance of 3 bits, given the disparity matching.  

\subsection{Planar Graph Optimization} \label{sec:stereo_optimization}
In the stereo factor graph, we only estimate the planes $\plane$ from the factors in Eq.~\ref{eq:census_factor}, i.e. we constrain the motion $\motion$ to be constant
(Eq. \ref{eq:matching_factor}, \ref{eq:smooth_boundary_plane}). Suppose for each Gaussian noise factor, $r$ is its residual: $f(x) = \exp(-r(x))$. We can obtain the maximum a posterior (MAP) of the factor graph by minimizing the residuals in the least-square problem: 

\begin{equation} \label{eq:factor_graph_stereo}
\begin{aligned}
\plane^{\star} &= \argmax_{\plane} \prod f_{pho}(\plane_i) \cdot \prod f_{match}(\plane_i) \cdot \prod f_{smoothB}(\plane_i, \plane_j) \\
			&= \argmin_{\plane} \sum r_{pho}(\plane_i) + \sum r_{match}(\plane_i) + \sum r_{smoothB}(\plane_i, \plane_j)
\end{aligned}
\end{equation}

Levenberg-Marquardt can be used to solve this equation as a more robust choice (e.g. compared to Gauss-Newton), trading off efficiency for accuracy. 

\subsection{Semi-dense Matching \& Multi-Hypotheses RANSAC}\label{sec:multi_hypotheses_ransac}

\begin{figure}[t]
\begin{center}
\subfloat{\includegraphics[width=0.30\linewidth]{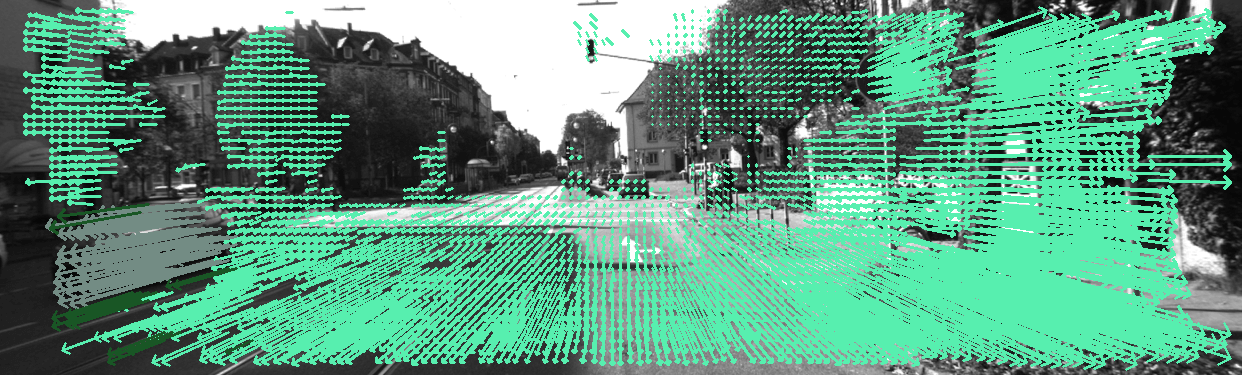}} \ \
\subfloat{\includegraphics[width=0.30\linewidth]{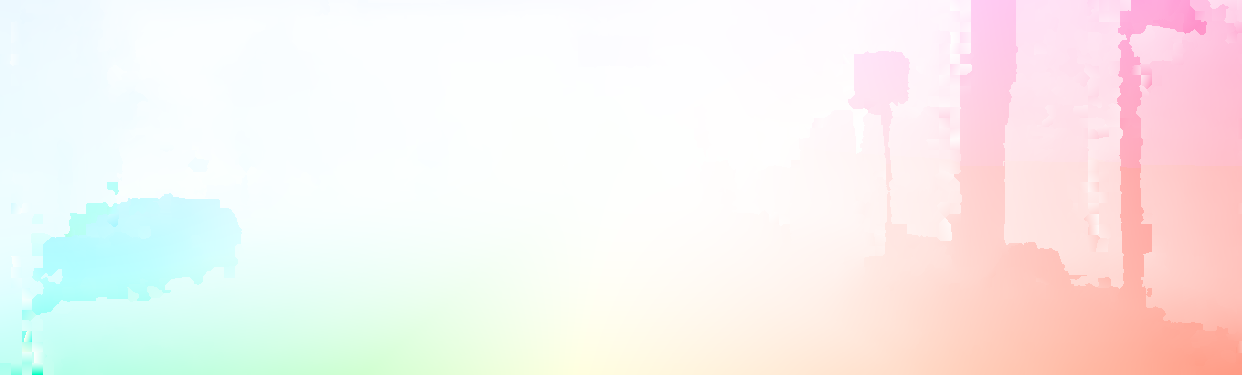}} \ \
\subfloat{\includegraphics[width=0.30\linewidth]{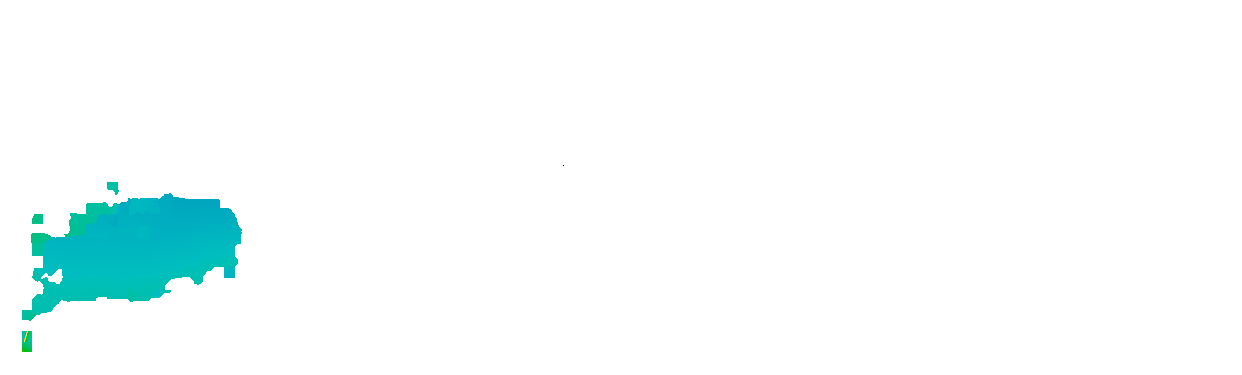}}   \\
\subfloat{\includegraphics[width=0.30\linewidth]{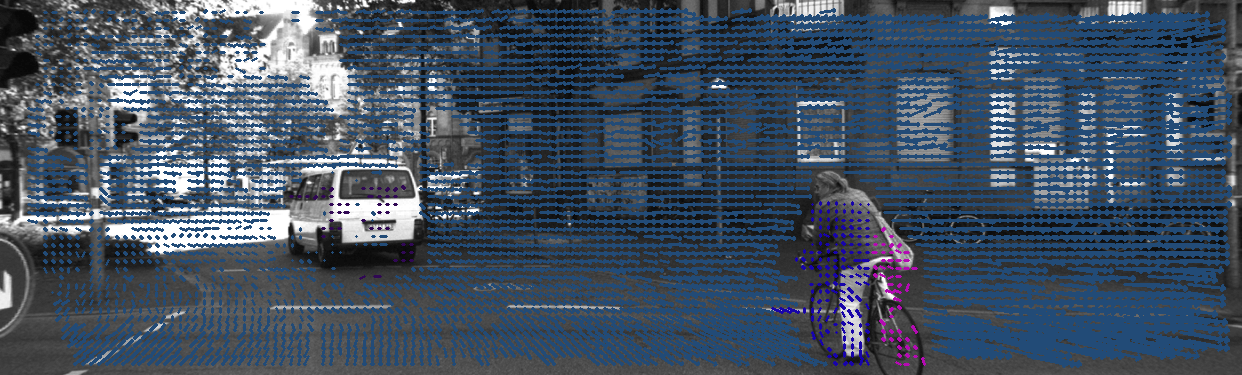}} \ \
\subfloat{\includegraphics[width=0.30\linewidth]{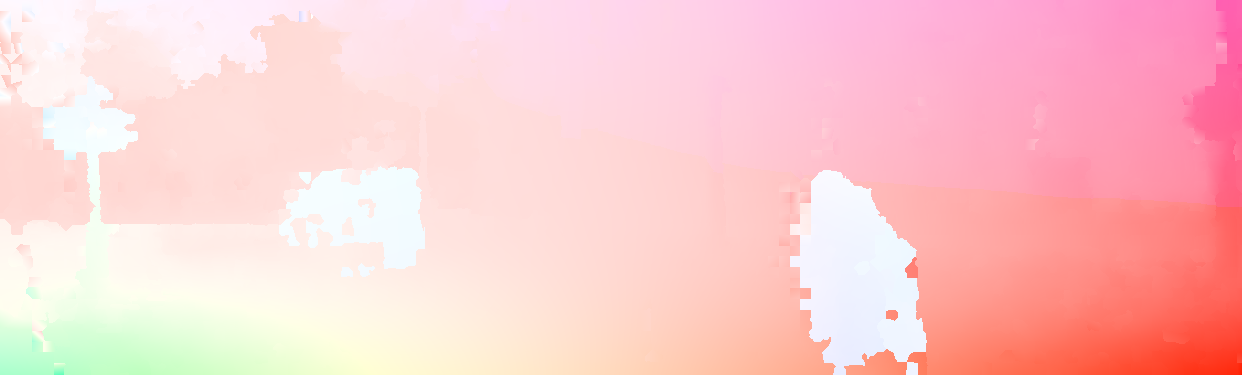}} \ \
\subfloat{\includegraphics[width=0.30\linewidth]{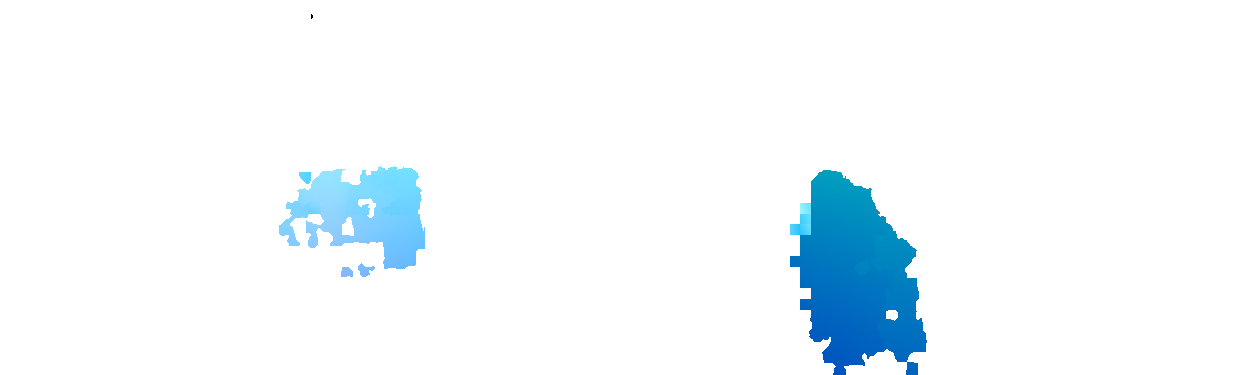}}
\caption{A visualization of motion hypothesis (left), optical flow (middle), and scene motion flow (right). Camera motion is explicitly 
removed from scene motion flow. In the image of the cyclist we show that although multiple motion hypotheses are discovered by RANSAC (in two 
colors), a final smooth motion over this non-rigid entity is estimated with continuous optimization.}
\label{fig:motion_putatives}
\end{center}
\end{figure}

We leverage the state-of-art matching method~\cite{Weinzaepfel13iccv} to generate a semi-dense matching field, which has the advantage of being able to associate across large displacements in the image space. To estimate the initial motion for superpixels, we chose RANSAC similar to \cite{Menze15cvpr}. We classify putatives as inliers based on their re-projection errors. The standard-deviation $\sigma = 1$ is small to ensure that bad hypotheses are rare. All hypotheses with more than $20\%$ inliers in each step are retained. Compared to the up-to-5 hypotheses in \cite{Menze15cvpr}, we found empirically that our RANSAC strategy can retrieve 10-20 hypotheses in complex scenes, which ensures a high recall of even small moving objects, or motion patterns on non-rigid objects (e.g. pedestrians and cyclists). This process can be quite slow when noisy matches are prominent and inliers ratios are low. To cope with this effect, we use superpixels as a prior in RANSAC. We evaluate the inlier superpixels (indicated by inlier feature matches through non-maximum suppression), and reject conflicting feature matches as outliers. This prunes the number of motion hypotheses, and substantially speeds up this step. See Figure \ref{fig:motion_putatives} for an illustration of the motion hypotheses.

Since the most dominant transform in the scene is induced by the camera transform, we can get an estimate of the incremental camera transform in the first iteration. After each iteration, the hypothesis is refined by a weighted least squares optimization, solved efficiently by Levenberg-Marquardt. 

\subsection{Local Motion Estimation} \label{sec:local_motion_estimation}
After estimation of the plane itself, we initialize the motion $\motion_i$ of each individual plane from the set of motion hypotheses. At this step, given the raw 
image measurements $I_{0,1}$, a pair of  estimated depth maps in both frames $D_{0,1}$, and the sparse point-matching field $F$, the goal is to 
estimate the most probable hypothesis $l^{\star}$  for each individual superpixel. We assume a set of conditional independencies among $I_{0,1}$, $D_{0,1}$, and $F$, given the superpixel. The label $l$ for each superpixel can therefore be inferred from the Bayes rule:

\begin{equation}
\begin{aligned}
P(l | F, I_{0,1}, D_{0,1}) &\propto P (F, I_{0,1}, D_{0,1}| l)P(l)  \\ &\propto P(I_{0,1}|l) P(D_{0,1}|l) P(F, I_0, D_0 |l) P(l),
\end{aligned}
\end{equation}

Assuming each motion hypothesis has equally prior information, a corresponding MAP estimation to the above equation can be presented as:

\begin{equation}
l^{\star} = \argmax_{l^{\star}} \mathbf{E}_{depth}(l) + \alpha \mathbf{E}_{photometric}(l) + \beta \mathbf{E}_{cluster}(l),
\end{equation}

\noindent where $\mathbf{E}_{depth}(l)$ represents the depth error between the warped depth and transformed depth, given a superpixel and its plane; $\mathbf{E}_{photometric}(l)$ represents the photometric error between the superpixel and its warped 
superpixel; $\mathbf{E}_{cluster}(l)$ represents the clustering error of a superpixel, w.r.t. its neighborhood features:  

\begin{equation} 
\begin{aligned}
\mathbf{E}_{depth}(l) &= \sum_{p_i \in S} (D_1(\mathbf{H} p_i)) - z(\mathbf{H} p_i))^2, \\
\mathbf{E}_{photometric}(l) &= \sum_{p_i \in S} (I(p_i) - I(\mathbf{H}p_i))^2, \\
\mathbf{E}_{cluster}(l) &= \sum_{p_i \in S} \sum_{p_k \in F_l} \exp(-\frac{\bigtriangledown I_{i,k}^2}{\sigma^2_i}) 
\exp(-\frac{\bigtriangledown D_{i,k}^2}{\sigma^2_{D}}),
\end{aligned}
\end{equation}

\noindent where $\mathbf{H}$ is the homography transform and $z(p)$ is the depth at pixel $p$. $\bigtriangledown I_{i,k}^2$ and $\bigtriangledown D_{i,k}^2$ describes the color and depth difference of a pixel $p_i \in S$ to a 
feature point $p_k \in F_l$ belonging to hypothesis $l$. $\sigma_I$ and $\sigma_D$ are their variances.

A local motion optimization is done for each hypothesis by incorporating the factors \ref{eq:census_factor}, \ref{eq:matching_factor}, 
\ref{eq:smooth_boundary_motion}, \ref{eq:smooth_motion} with pre-estimated planes values as:

\begin{equation}
\begin{aligned}
\motion^{\star} = \argmin_{\motion} &\sum r_{pho}(\motion_i) + \sum r_{match}(\motion_i) + \sum r_{smoothB}(\motion_i, \motion_j) + \\
&\sum r_{smoothM}(\motion_i, \motion_j) + \sum r_{prior}(\mathcal{M}).
\end{aligned}
\end{equation}

Similar to Eq.~\ref{eq:factor_graph_stereo}, $r$ is the residual for each factor. We add a prior factor $f_{prior}(\cdot)$ to enforce an $L_2$ prior centered at $0$. It works as a diagonal term to improve the condition numbers in the matrix factorization. The prior factor has small weights and in general do not affect the accuracy or speed significantly.

\subsection{Global Optimization}\label{sec:global_optimization}
Finally, we estimate the global factor graph, with the complete set of parameters $\planes = \{\plane, \motion\}$ in the reference frame. 
The factors in this stage are set using measurements in all of the other three views, w.r.t. reference image. 

\begin{equation}
\begin{aligned}
\planes^{\star} = \argmin_{\planes} &\sum r_{pho}(\planes_i) + \sum r_{match}(\planes_i) + \sum r_{smoothB}(\planes_i, \planes_j) + \\
&\sum r_{smoothM}(\planes_i, \planes_j) + \sum r_{prior}(\planes_i)
\end{aligned}
\end{equation}
\section{Experiments and Evaluations}\label{sec:experiments}
Our factors and optimization algorithm are implemented using GTSAM \cite{Dellaert12tr}. As input to our method, we use super-pixels generated from 
\cite{Zitnick14eccv}, a fast stereo prior from \cite{Yamaguchi14eccv}, and the DeepMatching method in~\cite{Weinzaepfel13iccv}. 
The noise models and robust kernel thresholds of the Gaussian factors are selected based on the first 100 training images in KITTI. In the next subsections, we discuss the results as well as optimization and individual factor contribution to the results. 

\subsection{Evaluation Over KITTI} \label{sec: kitti_evaluation}
\begin{figure}[t] 
\begin{center}
\tiny
\begin{tabular}{cccc}
\multicolumn{4}{c}{raw images in reference view} \\
\includegraphics[width=0.24\linewidth]{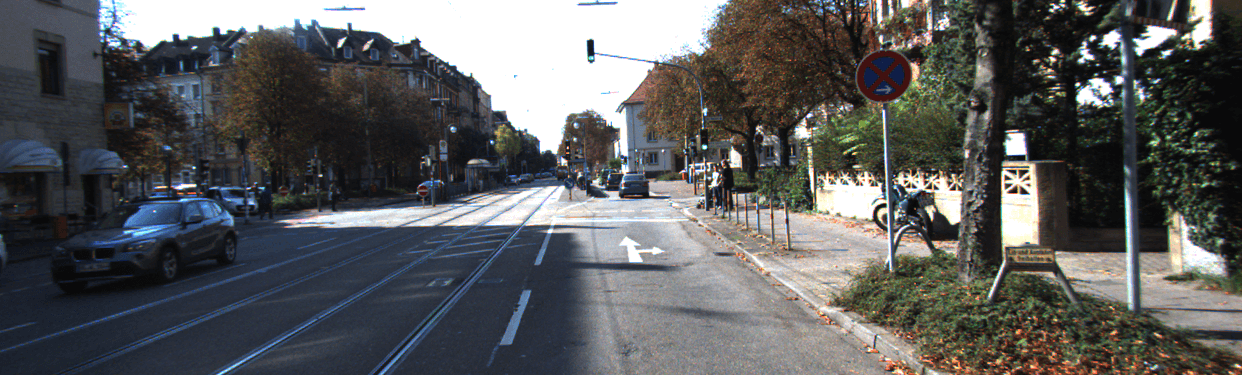} &
\includegraphics[width=0.24\linewidth]{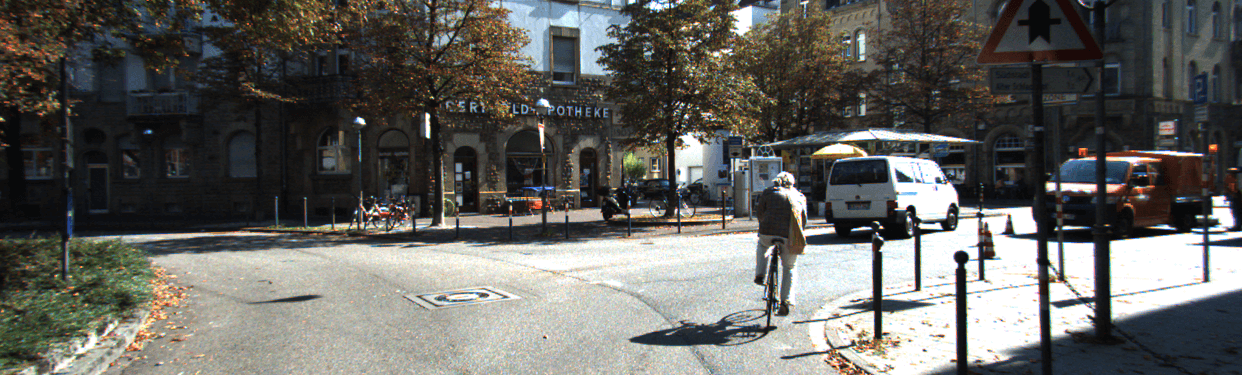} &
\includegraphics[width=0.24\linewidth]{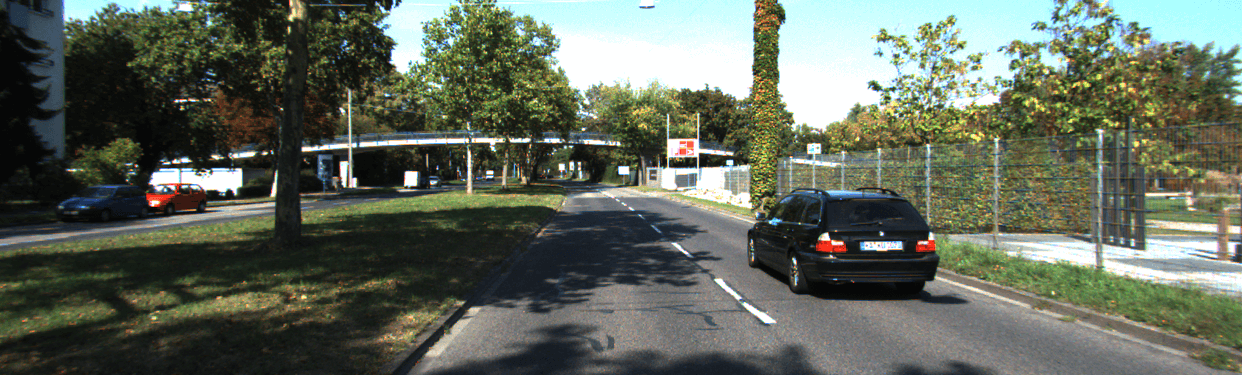} &
\includegraphics[width=0.24\linewidth]{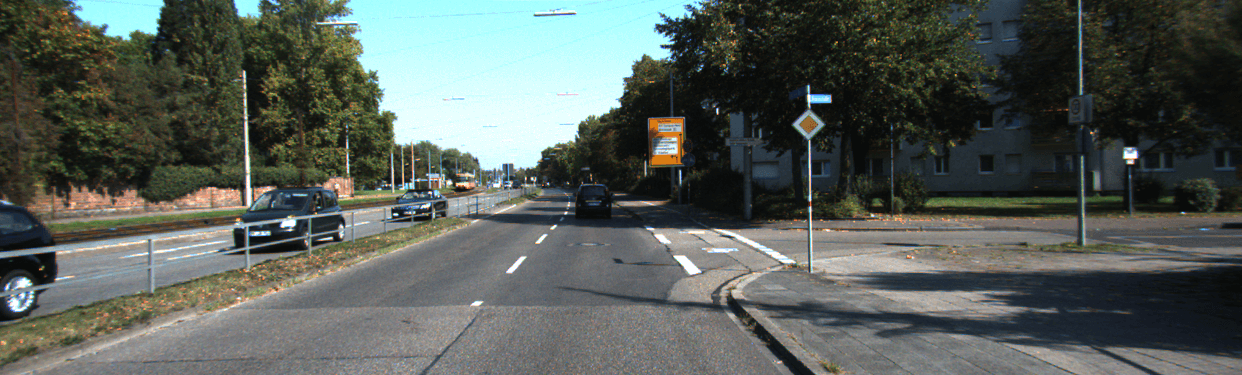} \\

\hline\noalign{\smallskip}
\multicolumn{4}{c}{estimated disparity (top), ground truth (middle), error (down) } \\
\includegraphics[width=0.24\linewidth]{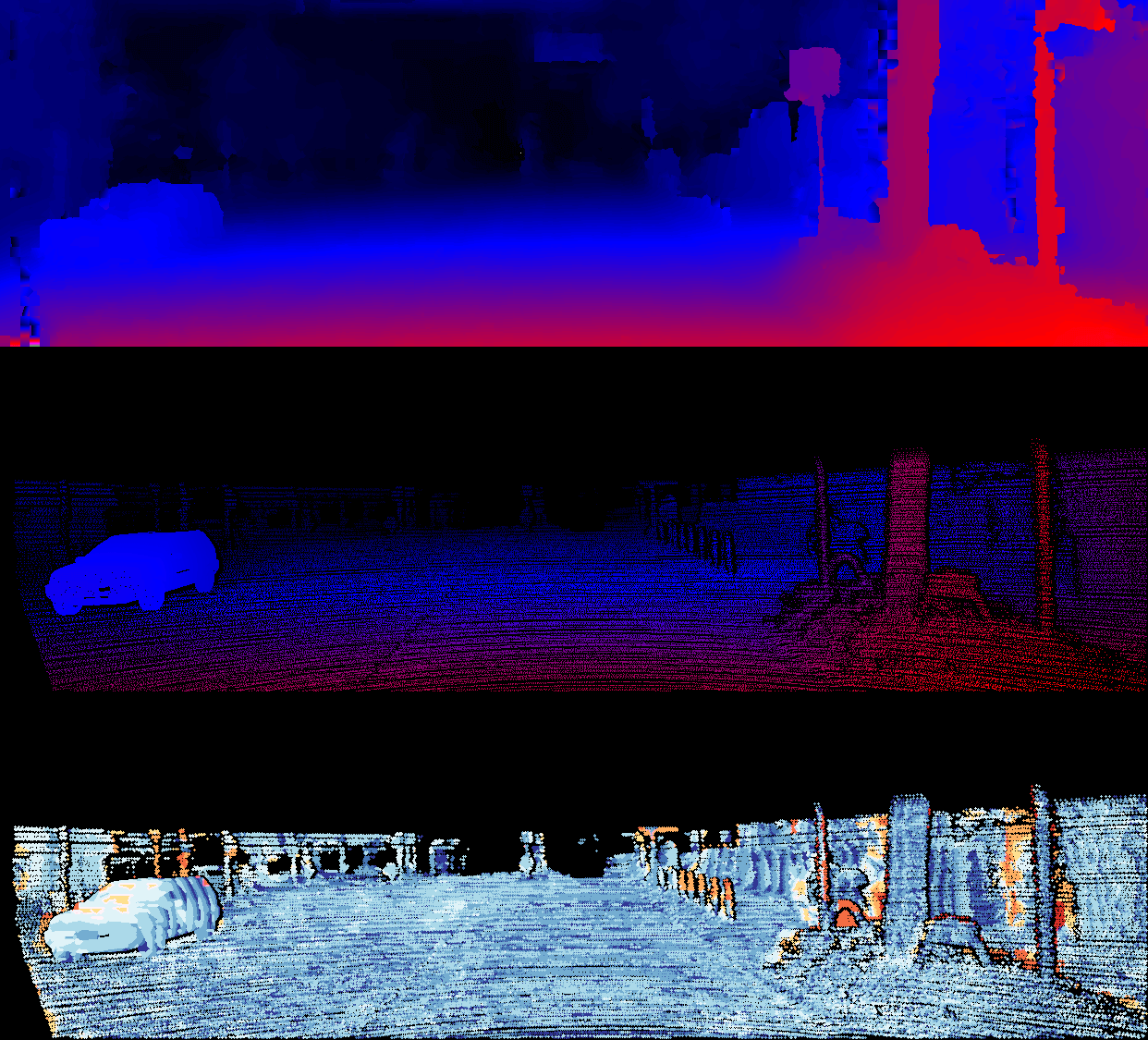} &
\includegraphics[width=0.24\linewidth]{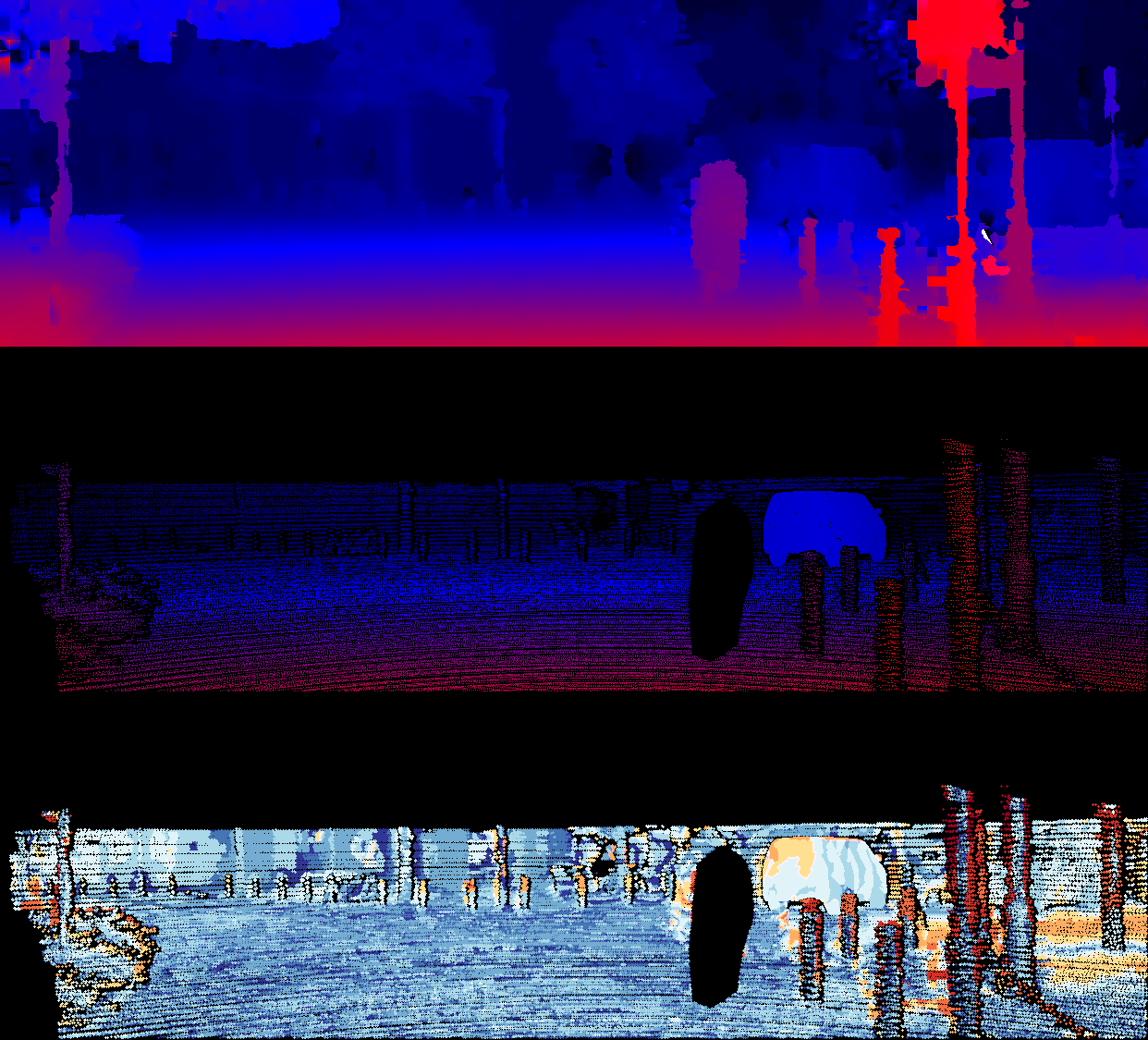} &
\includegraphics[width=0.24\linewidth]{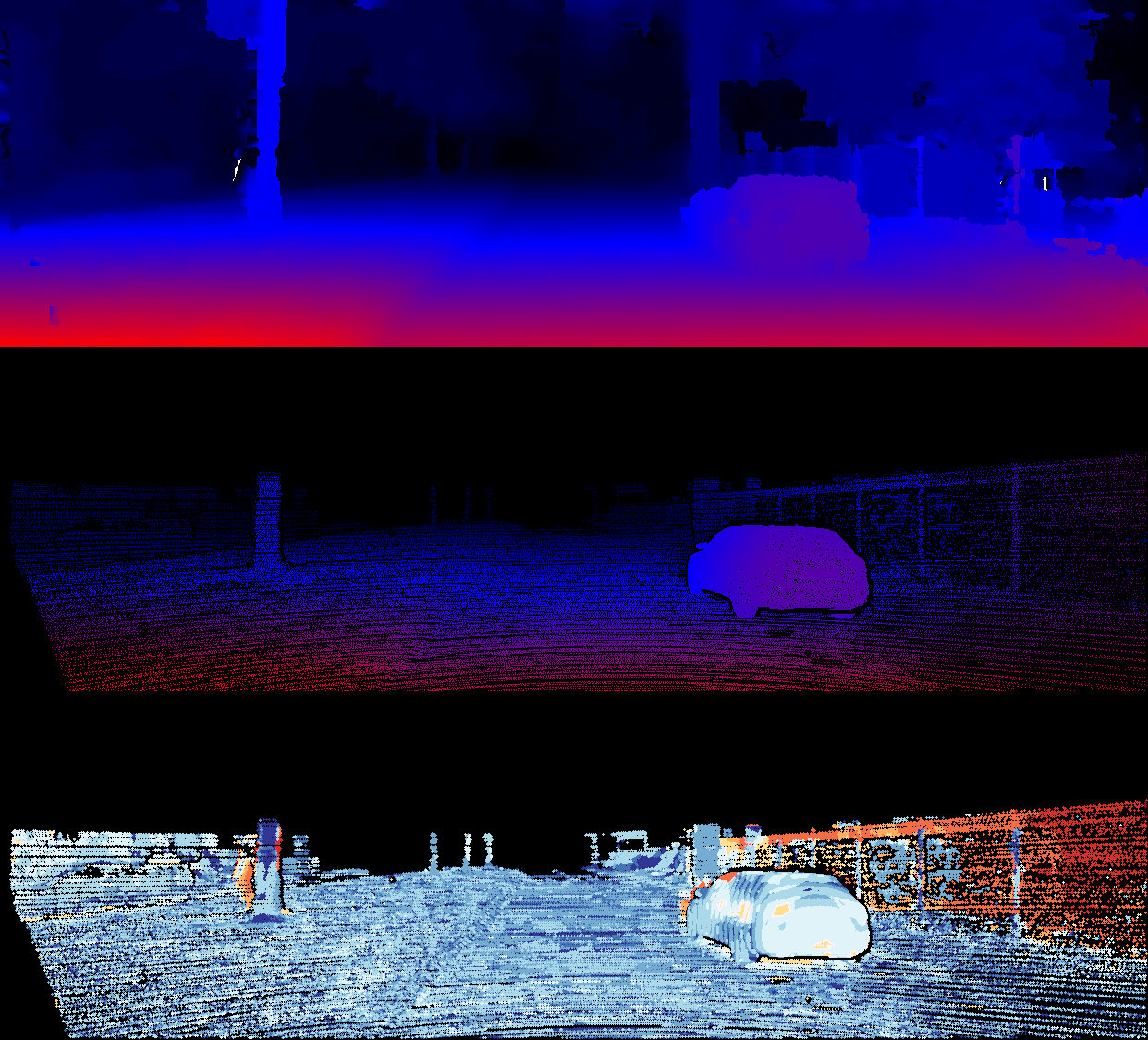} &
\includegraphics[width=0.24\linewidth]{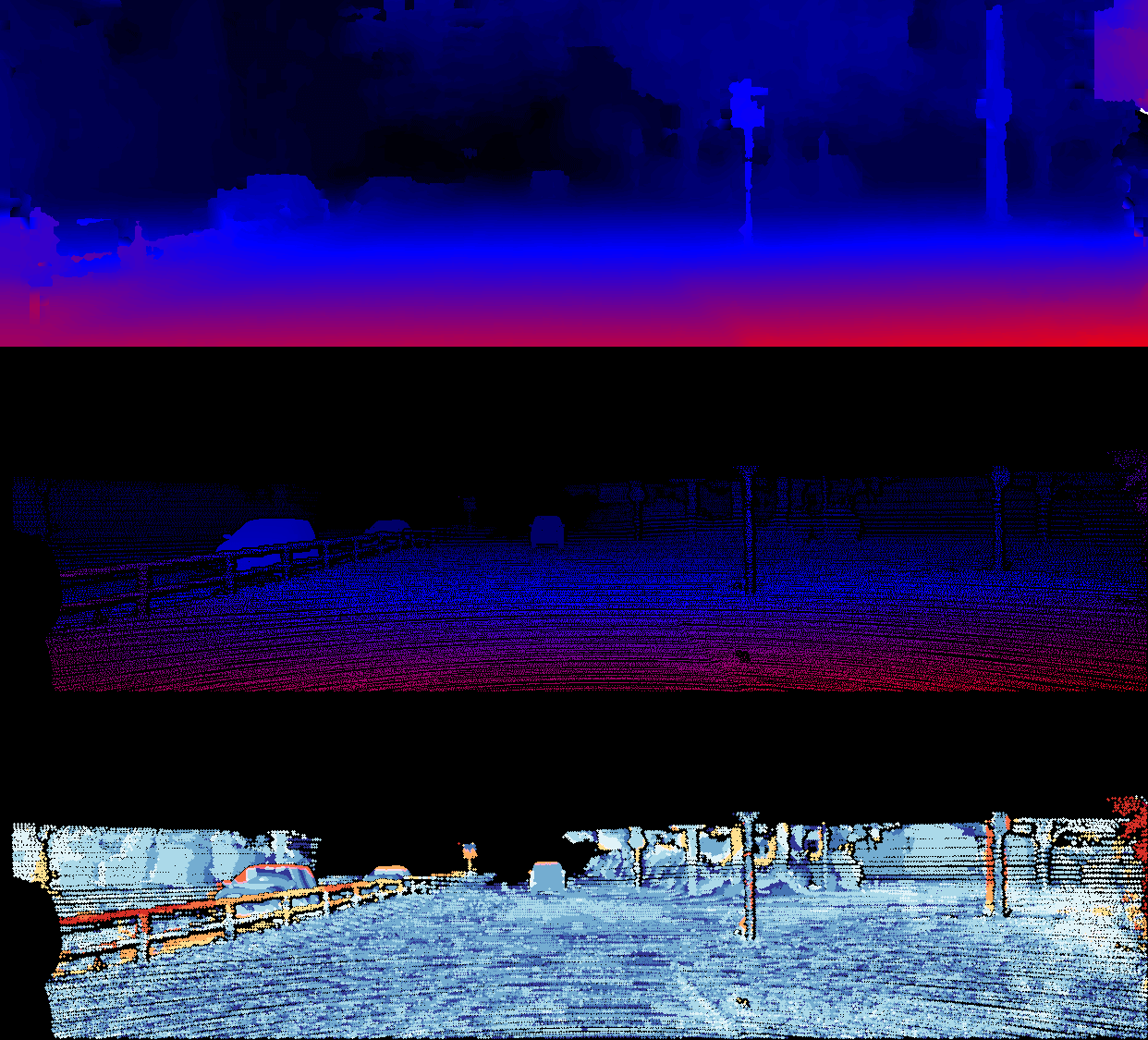} \\

\hline\noalign{\smallskip}
\multicolumn{4}{c}{estimated flow (top), ground truth (middle), error (down) } \\
\includegraphics[width=0.24\linewidth]{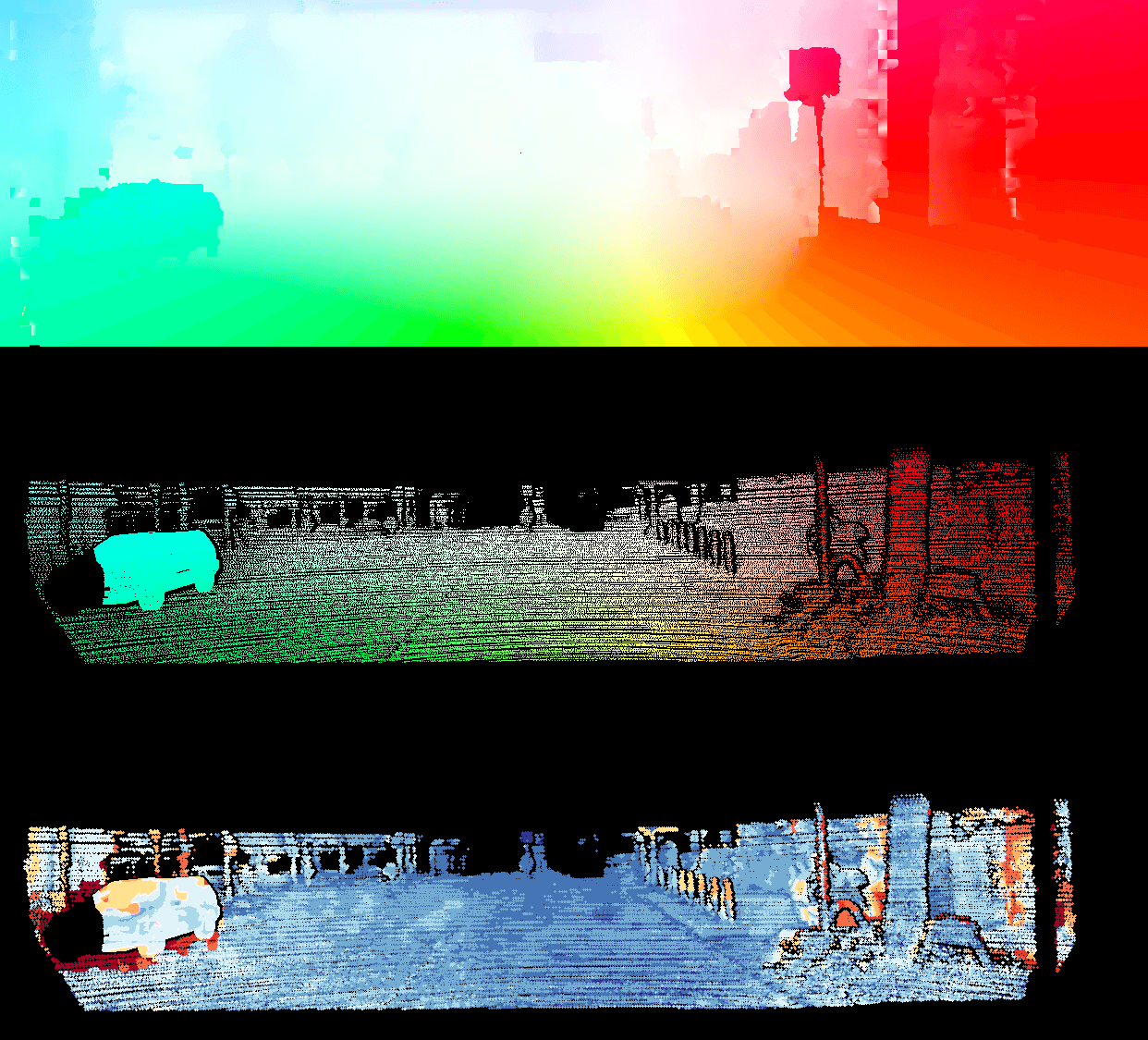} &
\includegraphics[width=0.24\linewidth]{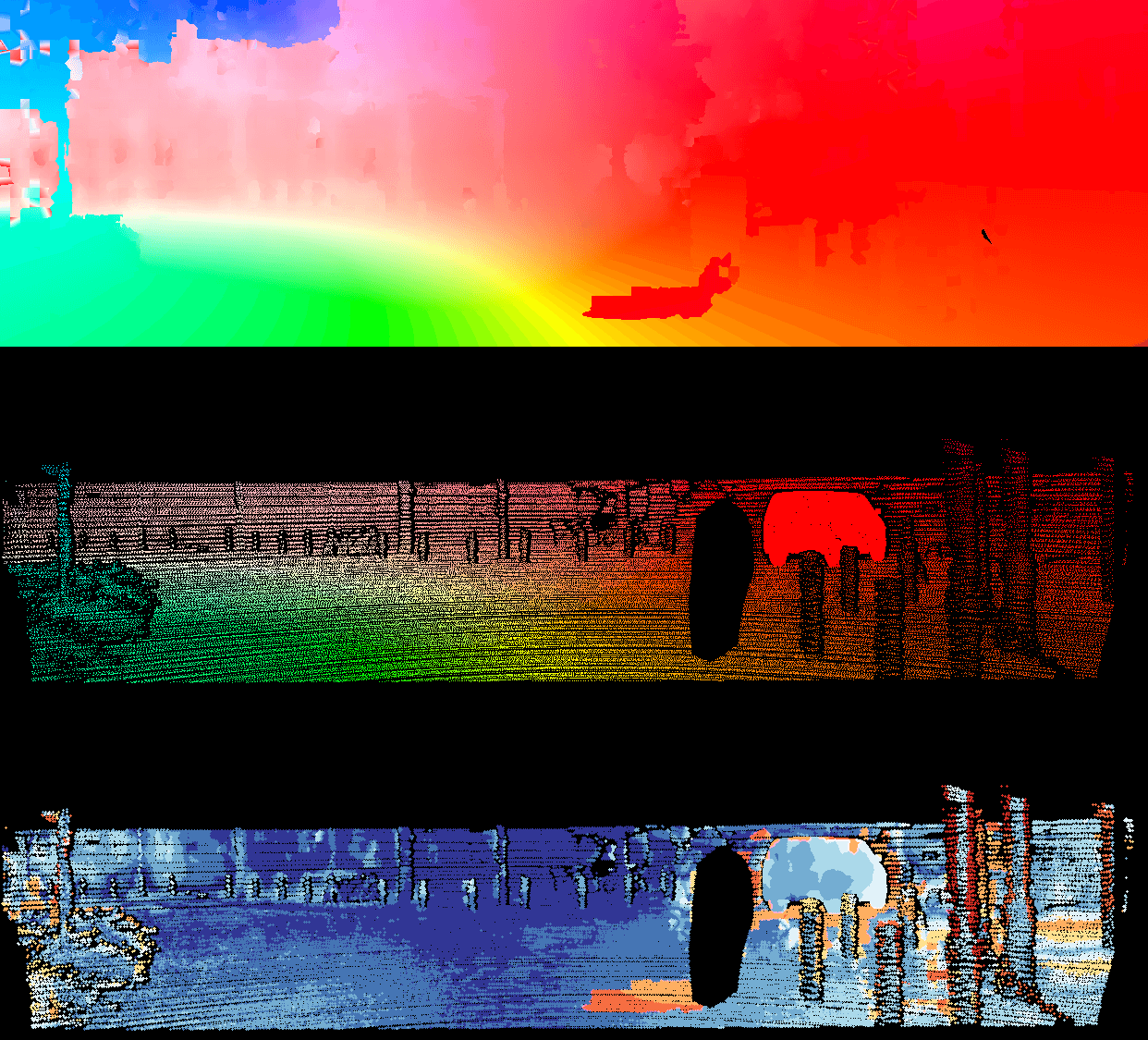} &
\includegraphics[width=0.24\linewidth]{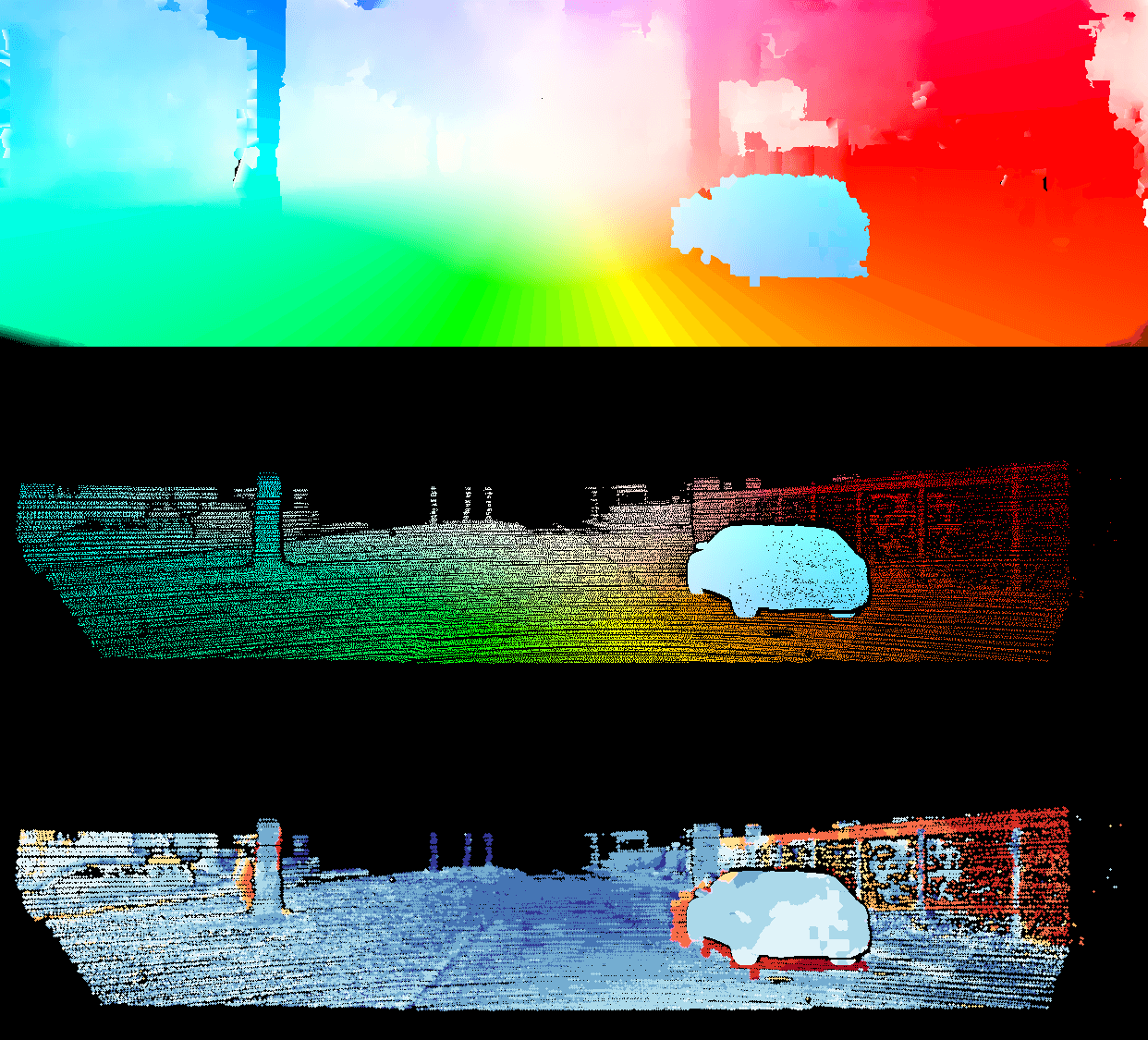} &
\includegraphics[width=0.24\linewidth]{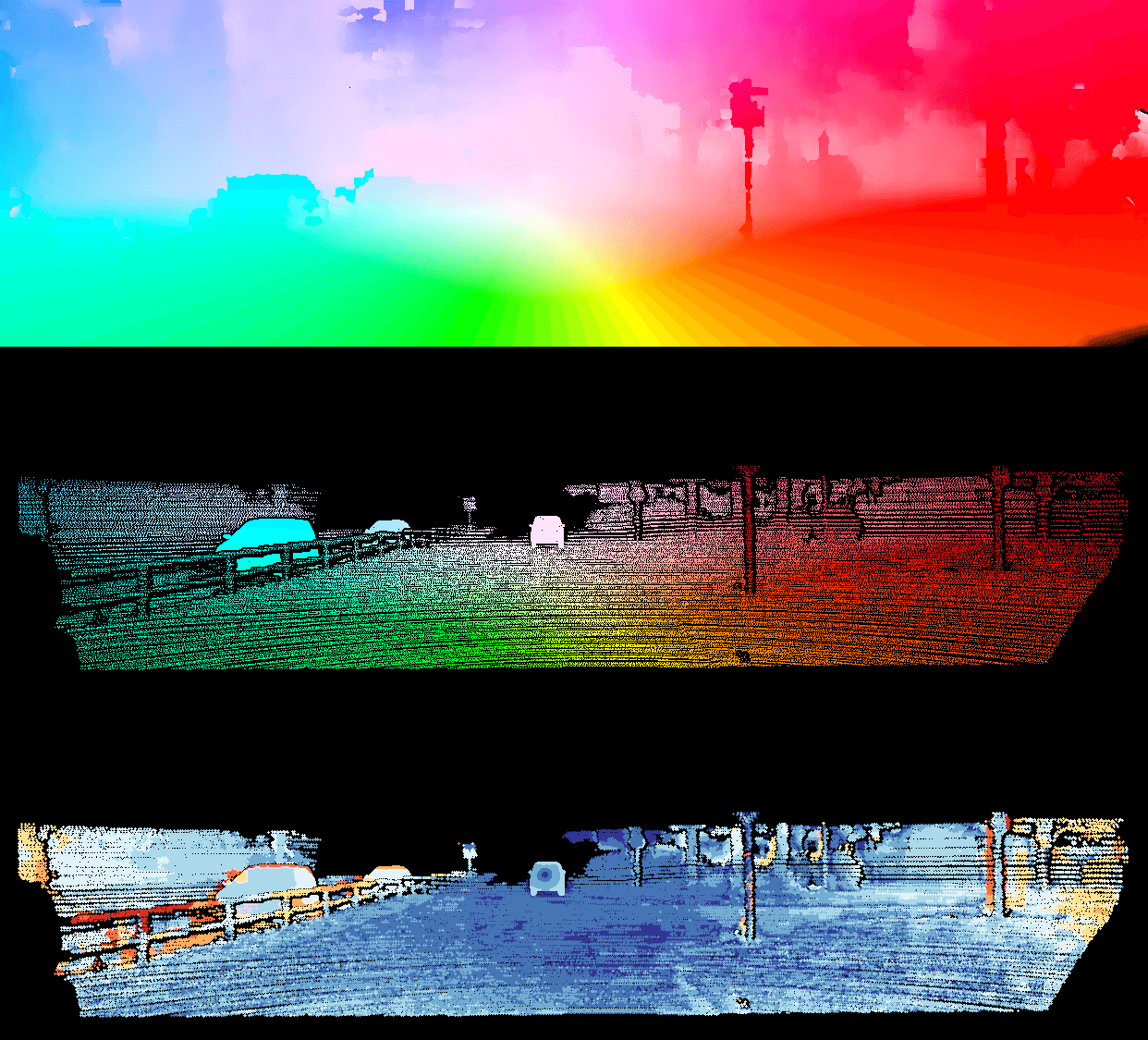} \\

\end{tabular}
\end{center}

\caption{\textbf{Qualitative Results in KITTI.} We show the disparity and flow estimation against the ground truth results in Kitti Scene Flow training set. }
\label{fig:qualitative_evaluation}
\end{figure}

We evaluate our algorithm on the challenging KITTI Scene Flow benchmark~\cite{Menze15cvpr}, which is a realistic benchmark in outdoor 
environments. In the KITTI benchmark, our method ranks \emph{3rd in Scene Flow test} while being significantly faster than close competitors, as well as \emph{3nd in the KITTI Optical Flow test} and 11th in 
the stereo test which we did not explicitly target. We show our quantitative scene flow results in Table~\ref{tab: kitti_scene_flow_evaluation} and 
qualitative visualizations in Fig.~\ref{fig:qualitative_evaluation}. 

\begin{table}[t]
\tiny
\captionsetup{font=scriptsize}
\caption{\textbf{Quantitative Results on KITTI Scene Flow Test Benchmark}. We show the disparity errors reference frame (D1) and second frame (D2), flow error (Fl), and the scene flow (SF) in 200 test images on KITTI. The errors are reported as background (bg), foregound (fg), 
and all pixels (bg+fg), OCC for errors over all areas, NOC only for errors non-occluded areas.}
\label{tab: kitti_scene_flow_evaluation}

\begin{center}

\begin{tabular}{c|c|c|c|c|c|c|c|c|c|c|c|c|c}
\hline\noalign{\smallskip}
\multirow{3}{*}{Method} & \multicolumn{13}{c}{Occlusion (OCC) error}								  \\
\hhline{~}
						&\multicolumn{3}{c}{D1}   &\multicolumn{3}{c}{D2}	& \multicolumn{3}{c}{Fl} &  
\multicolumn{3}{c}{SF} &   \\  
\hhline{~}
 				 		& bg\%  & fg\%   & all\%  & bg\% & fg\% & all\%  	& bg\% & fg\% & all\%    & bg\%  & 
fg\% & all\%   & time	\\
\hline
PRSM\cite{Vogel15ijcv} & \textbf{3.02}&\textbf{10.52}&\textbf{4.27}&\textbf{5.13}&\textbf{15.11}&\textbf{6.79}&\textbf{5.33}&
\textbf{17.02}&\textbf{7.28}&\textbf{6.61}&\textbf{23.60}&\textbf{9.44} & 300 s \\

OSF \cite{Menze15cvpr}& 4.54 & 12.03 & 5.79 & 5.45 & 19.41 & 7.77 	& 
 5.62 & 22.17 & 8.37 & 7.01 & 28.76 & 10.63  	& 50 min \\
PRSF \cite{Vogel13iccv} & 4.74  & 13.74  & 6.24 & 11.14 & 20.47 & 12.69	& 11.73 & 27.73 & 14.39  & 13.49 & 33.72 & 16.85  	& 150 s  \\
SGM+SF \cite{Hornacek14cvpr}&5.15 & 15.29& 6.84 & 14.10 & 23.13 & 15.60  	& 20.91 & 28.90 & 22.24  & 23.09 & 37.12 & 25.43 	& 45 
min  \\
SGM+C+NL\cite{Sun14ijcv}& 5.15  & 15.29  & 6.84 & 28.77 & 25.65 & 28.25  	& 34.24 & 45.40 & 36.10  & 38.21 & 53.04 & 40.68	& 
4.5 min  \\ 
VSF \cite{Huguet07iccv} & 27.73 & 21.72  & 26.38& 59.51 & 44.93 & 57.08  	& 50.06 & 47.57 & 49.64  & 67.69 & 64.03 & 67.08 	& 
125 min \\  
\hline
\textbf{Ours}      		& 4.57   & 13.04  & 5.98 & 7.92  & 20.76 & 10.06   & 10.40 & 30.33 & 13.71 & 12.21 & 36.97 & 16.33  & 
\textbf{80 s}    \\
\hline
\end{tabular}

\begin{tabular}{c|c|c|c|c|c|c|c|c|c|c|c|c|c}
\hline\noalign{\smallskip}
\multirow{3}{*}{Method} & \multicolumn{13}{c}{Non-Occlusion (NOC) error}								  \\
\hhline{~}
						 & \multicolumn{3}{c}{D1}	& \multicolumn{3}{c}{D2}& \multicolumn{3}{c}{Fl} &  
\multicolumn{3}{c}{SF} &   \\  
\hhline{~}
 				 		& bg\%  & fg\%   & all\%  	& bg\% & fg\% 	& all\%  & bg\% & fg\% & all      &  bg\% & 
fg\% & all    & time	\\
\hline
PRSM\cite{Vogel15ijcv} & \textbf{2.93}&\textbf{10.00}&\textbf{4.10}&\textbf{4.13}&\textbf{12.85}&\textbf{5.69}& 4.33 &
\textbf{14.15}&\textbf{6.11}&\textbf{5.54}&\textbf{20.16}&\textbf{8.16} & 300 s \\

OSF \cite{Menze15cvpr}  & 4.14  & 11.12 & 5.29 & 4.49 & 16.33 	& 6.61   & \textbf{4.21}  & 18.65 & 6.83   & 5.52  & 24.58 & 8.93  	
& 50 min \\
PRSF \cite{Vogel13iccv} & 4.41  & 13.09  & 5.84 	& 6.35 & 16.12 	& 8.10   & 6.94  & 23.64 & 9.97   & 8.35  & 28.45 & 11.95 & 150 s  
\\
SGM+SF \cite{Hornacek14cvpr}&4.75&14.22  & 6.31 	& 8.34 & 18.71 & 10.20  & 13.36 & 25.21 & 15.51  & 15.28 & 32.33 & 18.33 	& 45 
min  \\
SGM+C+NL\cite{Sun14ijcv}& 4.75 & 14.22& 6.31 		& 15.72& 20.79 & 16.63  & 23.03 & 41.92 & 26.46  & 26.22 & 48.61 & 30.23	& 
4.5 min  \\ 
VSF \cite{Huguet07iccv} & 26.38 & 19.88  & 25.31	& 52.30& 40.83 & 50.24  & 41.15 & 44.16 & 41.70  & 61.14 & 60.38 & 61.00 	& 
125 min \\  
\hline
\textbf{Ours}      		& \textbf{4.03}& 11.82  & 5.32   & 6.39  & 16.75 & 8.25   & 8.72 & 26.98 & 12.03  & 10.26 & 32.58 & 14.26  & 
\textbf{80 s}    \\
\hline
\end{tabular}

\end{center}
\end{table}

\begin{figure}[h!]
\begin{center}
\subfloat{\includegraphics[width=0.45\linewidth]{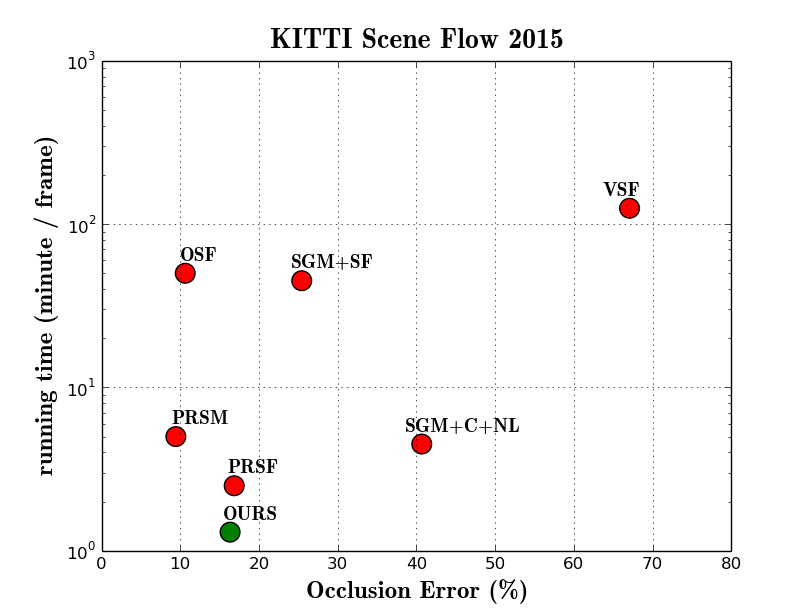}} 
\subfloat{\includegraphics[width=0.45\linewidth]{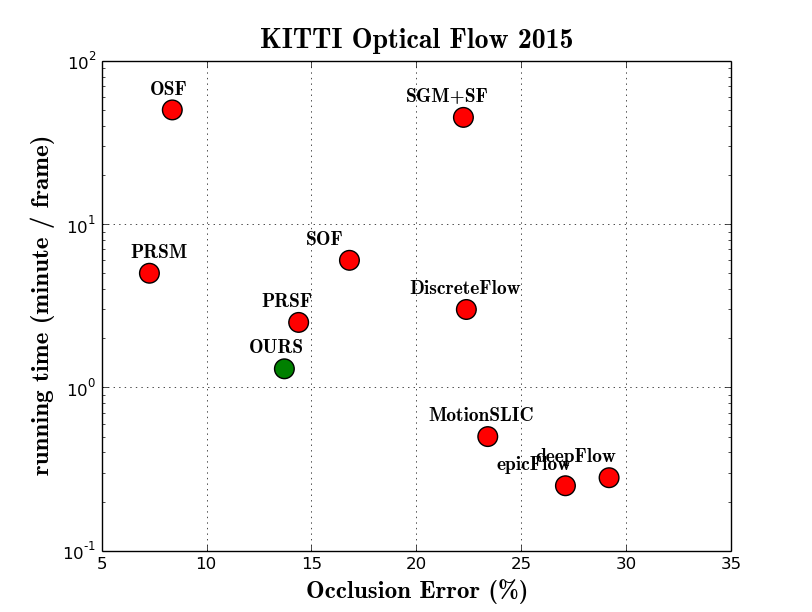}} 
\caption{Occlusion error-vs-time on KITTI. The running time axis is plotted in log scale. Our method is highlighted as green, which achieves top performance both in accuracy and computation speed.}
\label{fig:error_vs_time}
\end{center}
\end{figure}

Table~\ref{tab: kitti_scene_flow_evaluation} shows a comparison of our results against the other top 4 publicly-evaluated scene flow 
algorithms. 
In addition, we also added \cite{Huguet07iccv} (which proposed the four-image setting in scene flow) as a general comparison. In all of these 
results, the errors in disparity and flow evaluation are counted if the disparity or flow estimation exceeds 3 pixels and 5\% of its true 
value. In the Scene Flow evaluation,  the error is counted if any pixel in any of the three estimates (two stereo frame disparity images and 
flow image) exceed the criterion. We plot a error-vs-time figure in Fig. \ref{fig:error_vs_time}, which shows that our method achieves state-of-art performance, when considering both efficiency and accuracy.

Our results show a small difference in occlusion-errors, although occlusion is not directly handled as discrete labels. We follow the same 
representation in~\cite{Vogel13iccv} and achieved better performance in overall pixel errors and faster inference. Compared to all of these 
methods, our method is the fastest. Detailed test results are presented in our supplementary materials.

\begin{table}[t] 
\caption{\textbf{Quantitative Results on KITTI Optical Flow 2015 Dataset}. The errors are reported as 
background error(Fl-bg), foreground error (Fl-fg), and all pixels (Fl-bg+Fl-fg), NOC for 
non-occluded areas error and OCC for errors over all pixels. Methods that use stereo information are shown as \textit{italic}. 
}
\label{tab:kitti_optical_flow_evaluation}
\begin{center}
\scriptsize
\begin{tabular}{c|c|c|c|c|c|c|c}
\hline\noalign{\smallskip}
\multirow{2}{*}{Method} & \multicolumn{3}{c}{ OCC error} & \multicolumn{3}{c}{NOC error} &  \\
\hhline{~}

						 &  Fl-bg\%  & Fl-fg\%   & all\%  	& Fl-bg\% & Fl-fg\% 	& all\% &  time \\
\hline
\textit{PRSM}\cite{Vogel15ijcv}&\textbf{5.33}&\textbf{17.02}&\textbf{7.28}& 4.33 & \textbf{14.15} & \textbf{6.11} & 300 s \\

\textit{OSF}\cite{Menze15cvpr} & 5.62 & 22.17 & 8.37 & \textbf{4.21} & 18.65 & 6.83 & 50 min\\
\textit{PRSF}\cite{Vogel13iccv} & 11.73 & 27.32 & 14.39 & 6.94 & 23.64 & 9.97  &  150s\\
SOF \cite{Sevilla16cvpr} & 14.63 & 27.73 & 16.81 & 8.11 & 23.28 & 10.86  & 6 min\\
\textit{SGM SF}\cite{Hornacek14cvpr} &  20.91 & 28.90 & 22.24 & 13.36 & 25.21 & 15.51 & 45 min\\
DiscreteFlow\cite{Menze15gcpr} &  21.53 & 26.68 & 22.38 &  9.96 & 22.17 & 12.18 & 3 min \\
\textit{MotionSLIC} \cite{Yamaguchi14eccv} & 14.86 & 66.21 & 23.40 & 6.19 & 64.82 & 16.83 & 30s \\
epicFlow \cite{Revaud15cvpr} & 25.81 & 33.56 & 27.10 & 15.00 & 29.39 & 17.61 & \textbf{15s} \\
deepFlow \cite{Weinzaepfel13iccv} & 27.96 & 35.28 & 29.18 & 16.47 & 31.25 & 19.15 & 17s \\
\hline
\textit{\textbf{ours}} 		&	10.40 & 30.33 & 13.71 & 8.72 & 26.98 & 12.03 & 80 s \\
\hline
\end{tabular}

\end{center}
\end{table}

Table \ref{tab:kitti_optical_flow_evaluation} shows our method compared to state-of-art optical flow methods. Methods using stereo information are 
shown in italic. The deepFlow \cite{Weinzaepfel13iccv} and epicFlow \cite{Revaud15cvpr} methods are also presented; these also leverage 
DeepMatching for data-association.   
Our method is third best for all-pixels estimation.

\subsection{Parameter Discussions} \label{sec:evaluate_factors}
In Table~\ref{tab:discuss_factors}, we evaluate the choice of each factor and their effects in the results. During motion estimation, we 
see that multi-scale Census has an important positive effect in improving convergence towards the optima. Note that the best choice of 
weights for each factor was tuned by using a similar analysis. A more detailed parameter analyses is presented in the supplement materials. 

\begin{table}[t] 
\begin{center}
\caption{Evaluation over factors. The non-occlusion error are used from 50 images of KITTI training set. The corresponding factors (in 
braces) are in section \ref{sec:factor_graph}}.
\label{tab:discuss_factors}
\tiny
\begin{tabular}{c|c|c|c|c|c|c|c}
\hline\noalign{\smallskip}
\multicolumn{4}{c}{Stereo Error \% (Noc)} & \multicolumn{4}{c}{Flow Error \% (Noc)} \\  
\hline\noalign{\smallskip}
Factors        & D1-bg \% 	& D1-fg \% & D1-all \%  &  Factors        & F-bg \% 	& F-fg \% & F-all \% \\
\hline
Census (\ref{eq:census_factor})    	& 9.21  & 19.22 	& 12.31  & Census Raw only (\ref{eq:census_factor}) & 10.9 & 34.25 & 14.20 \\                      
Matching (\ref{eq:matching_factor}) 	& 5.95	& 15.20		& 7.62  & Census Multi-scale (\ref{eq:census_factor}) & 9.3 & 30.13 & 12.45 \\
Census + Matching (\ref{eq:census_factor}, \ref{eq:matching_factor}) & 5.66 	 & 15.01  & 6.93 & Matching only (\ref{eq:matching_factor}) & 10.5 & 33.40 & 13.20  \\
Census + Continuity (\ref{eq:census_factor}, \ref{eq:smooth_boundary_plane}) & 4.85 	& 14.22  & 5.94 & Census + piecewise motion (\ref{eq:census_factor}, \ref{eq:smooth_motion}) & 9.0 & 29.01 & 12.45 \\
All (\ref{eq:census_factor}, \ref{eq:matching_factor}, \ref{eq:smooth_boundary_plane}) & 4.13 & 10.20 & 4.85  & Census + continuity (\ref{eq:census_factor}, \ref{eq:smooth_boundary_motion}) & 9.2 & 30.15 & 12.44 \\
	& & & &
All (\ref{eq:census_factor}, \ref{eq:matching_factor}, \ref{eq:smooth_motion},  \ref{eq:smooth_boundary_motion}) & 8.92 & 28.92 & 12.31  \\
\hline\noalign{\smallskip}
\end{tabular}
\begin{tabular}{c|c|c|c}
\hline

\end{tabular}

\end{center}
\end{table}

\section{Conclusions}\label{sec:conclusions}

We present an approach to solve the scene flow problem in continuous domain, resulting in a high accuracy (3rd) on the KITTI Scene Flow benchmark at a large computational speedup. We show that faster inference is achievable by rethinking the solution as a non-linear least-square problem, cast within a factor graph formulation. We then develop a novel initialization method, leveraging a multi-scale differentiable Census-based cost and DeepMatching. Given this initialization, we individually optimize geometry (stereo) and motion (optical flow) and then perform a global refinement using Levenberg-Marquardt. Analysis shows the positive effects of each of these contributions, ultimately leading to a fast and accurate scene flow estimation.

The proposed method has already achieved significant speed and accuracy, and several enhancements are possible. For example, there are several challenging points and failure cases that we do not cope with so far, such as photometric inconsistency in scenes and areas with aperture ambiguity. To address these problems, we expect to explore more invariant constraints than the current unary factors, and more prior knowledge to enforce better local consistency. Finally, it is possible that additional speed-ups could be achieved through profiling and optimization of the code. Such improvements in both accuracy and speed would enable a host of applications related to autonomous driving, where both are crucial factors. 

\textbf{Acknowledgment}. This work was supported by the National Science Foundation and National Robotics Initiative (grant \# IIS-1426998). Fuxin Li was partially supported by NSF \# 1320348.

\clearpage

\bibliographystyle{splncs}
\bibliography{refs,egbib}
\end{document}